\documentclass{article}
\usepackage{spconf,amsmath,graphicx,booktabs,gensymb,url}

\title{Joint calibration and mapping of satellite altimetry data using trainable variational models}

\name{Q. Febvre$^\star$, R. Fablet$^\star$, J. Le Sommer$^\dagger$, C. Ubelmann$^\ddagger$\thanks{This work was supported by CNES, CLS, ANR Melody Oceanix, IDRIS.}}
\address{$^\star$IMT Atlantique, UMR CNRS Lab-STICC, Brest FR \\ 
$^\dagger$Univ. Grenoble Alpes, CNRS, IRD, Grenoble, FR \\
$^\ddagger$OceanNext, La Terrasse, FR}
\begin{document}
%
\maketitle
\begin{abstract}
Satellite radar altimeters are a key source of observation of ocean surface dynamics.
However, current sensor technology and mapping techniques do not yet allow to systematically resolve scales smaller than 100km. 
With their new sensors, upcoming wide-swath altimeter missions such as SWOT should help resolve finer scales.
Current mapping techniques rely on the quality of the input data, which is why the raw data go through multiple preprocessing stages before being used. Those calibration stages are improved and refined over many years and represent a challenge when a new type of sensor start acquiring data.
Here we show how a data-driven variational data assimilation framework could be used to jointly learn a calibration operator and an interpolator from non-calibrated data .
The proposed framework significantly outperforms the operational state-of-the-art mapping pipeline and truly benefits from wide-swath data to resolve finer scales on the global map as well as in the SWOT sensor geometry.
\end{abstract}
\section{Introduction}

Sea surface dynamics play an important role in a wide set of problematics ranging from
climate modeling, maritime traffic routing, oil spill monitoring to marine ecology. On a global scale, sea surface currents are to a large extent retrieved from the mapping of sea surface height (SSH) fields using satellite nadir altimetry data \cite{rohrs2021}. 
As current nadir altimeter sensors involve a scarce and irregular space-time sampling of the ocean surface, the state-of-the-art mapping schemes fail to reconstruct scales lower than ~100km. In this context, the future wide-swath altimeter SWOT mission \cite{morrow2019} opens the perspective of being able to reconstruct finer scales.

Broadly speaking, the mapping of SSH fields from satellite altimetry data relies on two main steps: 
a calibration step to remove acquisition and geophysical noises and an interpolation step to produce gap-free maps from the irregularly-sampled calibrated observations. Among the different noise processes to be accounted for, we may cite both sensor noises, geometric noise patterns associated with random perturbations of the attitude of the satellite platform as well as geophysical processes which may superimpose to the SSH information \cite{swotsciencereq}.
The calibration step is paramount for current interpolation methods because they do not account for observation biases. We may categorize interpolation methods according to the prior they assume to represent sea surface dynamics. While the operational SOTA product \cite{duacs} exploits an optimal interpolation (OI) based on a covariance-based prior, assimilation-based products \cite{glorys} relies on the assimilation of ocean circulation models \cite{le2018ocean}.
Recently data-driven approaches, especially neural networks \cite{joint4dvar}, have shown some success for the interpolation of sea surface fields from satellite-derived observation data. As stated above, all these approaches may be strongly affected by observation biases and are likely poorly adapted to the processing of uncalibrated datasets. 

In this paper, we aim to extend the benefits of the data-driven philosophy to the whole satellite-derived SSH mapping task from calibration to interpolation. We expect such an end-to-end framework to allow for lighter processing chains and alleviate the cost of the manual calibration steps. 
We consider an inverse problem formulation such that the resulting interpolation explicitly accounts for noise patterns in the observation of the SSH from space.

Overall, our key contributions are as follows: (i) we introduce a novel physics-informed learning framework backed on a variational formulation for the joint calibration and mapping of satellite altimetry data; (ii) We demonstrate the robustness of the trained model w.r.t. SWOT acquisition noises; (iii) We show how the proposed framework is able to recover finer scales on both the global scale and on the along-track direction of the SWOT swath.


\section{PROBLEM STATEMENT AND RELATED WORK}

We formulate the SSH mapping and SWOT calibration problems as inverse problems. SSH mapping refers to the estimation of gap-free SSH fields on a regular grid from irregularly-sampled data, while SWOT calibration is the retrieval from raw measurements of the SSH on the wide-swath domain observed by SWOT sensor.

\subsection{SSH Mapping}
We classically formulate the estimation of a state $X$ from observation data $Y$, referred to in geoscience as data assimilation problem \cite{carrassi} ,  as the minimization of a variational cost:
\begin{align}
	\label{eqn:var}
	\hat{X} = \operatorname*{argmin}_X J(X, Y) + R(X)
\end{align}
where $J$ is the observation term, which evaluates the agreement between the state and the observed values, and $R$ a prior on the state.
When considering calibrated data, $J$ is a linear quadratic term associated with a Gaussian prior on the observation error. The SOTA operational method DUACS \cite{duacs} relies on an optimal interpolation, which defines $R$ using a covariance-based prior, {\em i.e.} $R(X) = X^{-1}W^{-1}X$ with $W$ the covariance matrix that captures the spatio-temporal structure of the SSH fields. With such linear-quadratic terms, minimization (\ref{eqn:var}) can be solved analytically \cite{10.1007/BFb0080117}.

In variational data assimilation (4DVar) schemes \cite{blum2009data}, the prior knowledge involves a dynamical model $\mathcal{M}$ which describes the time evolution of the state of interest as $dX/dt=\mathcal{M}(X(t))$. Term $R$ is then stated as
\begin{align}
	\label{eqn:4dprior}
	R(X) = ||X - \Phi(X)||
\end{align}
with $\Phi$ the flow operator, which solves a time integration problem:
\begin{align}
	\label{eqn:phi}
\Phi(X)(t) = x(t-\delta_t) + \int_{t-\delta_t}^t\mathcal{M}(X(u))du
\end{align}
where $\delta_t$ is the time step. A key issue in 4DVar frameworks is the parameterization of dynamical model $\mathcal{M}$. General ocean circulation models \cite{le2018ocean} lead to the inversion of the full ocean state dynamics, which may be highly-complex and unstable when considering only sea surface observation. One may also consider dynamical representation of sea surface dynamics. In this context, quasi-geostrophic priors \cite{dyninterp}\cite{bfn} are appealing but may be limited to specific dynamical regimes.  

Recently, deep learning schemes have also been investigated to solve inverse problems and design trainable variational priors \cite{lucas2018using}\cite{Mack_2020}\cite{Brajard_2020}. Especially, for application to sea surface dynamics, our recent works  \cite{benaichouche:hal-03139066,joint4dvar, e2e4dvar} support the relevance of such trainable formulations. 


\subsection{SWOT Calibration}

For space-borne earth observation, the calibration of raw satellite-derived data into geophysically-sound 
measurements is a critical issue \cite{ceoshandbook}. Referred as L1 and L2 products, calibration steps consist in separating the error from the targeted geophysical signal in the observed values. In the context of SWOT mission, proposed approaches exploit some prior knowledge onto geometry and spectral distribution of the different error sources\cite{clemcalib}\cite{swotcalib}, {\em e.g.} attitude and orbiting  uncertainties, impact of other geophysical processes, thermal noise... Here, we consider as baseline such an approach which also exploits the agreement between the gap-free SSH fields issued from nadir altimeter data and raw SWOT data.


\section{PROPOSED METHOD}

In this section We present the proposed physics-informed trainable variational framework for the joint calibration and mapping of satellite altimetry data,  referred to as 4DVarNet-CalMap.
We first walk through the variational formulation.
Then we will describe the learning scheme and finally we will go through some implementation details.


\subsection{VARIATIONAL FORMULATION}
We consider a multi-scale decomposition of the SSH to better account for the different data sources. Formally, let us denote by $X_{LR}$ the low-resolution component of the gap-free SSH fields over the entire domain of interest ${\cal{D}}$, by $X_{HR-map}$ the gap-free high-resolution SSH anomaly over ${\cal{D}}$, and by $X_{HR-cal}$ high-resolution SSH anomaly over SWOT swaths. 

Overall, we define as state $X$ the concatenation of these three components $X_{LR}, X_{HR-map}, X_{HR-cal}$ considered over a time window $\Delta T$. As such, the gap-free SSH fields over domain ${\cal{D}}$ is given by $ X_{map} = X_{LR} + X_{HR-map}$ and the  SSH fields restricted to SWOT swaths to $X_{cal} = X_{LR} + X_{HR-cal}$. As observation data, we assume to be provided with optimally-interpolated low-resolution fields $Y_{LR}$ and the aggregation of nadir and SWOT altimeter data $Y_{\textsc{SAT}}$. This leads to the following observation term $J(X,Y)$:
\begin{equation}
	\begin{split}
		J(X, Y) = & \lambda_1\left \|(X_{LR} - Y_{LR}) \right \|^2\\
	+ & \lambda_2\left \| (X_{LR} + X_{HR-map} - Y_{\textsc{SAT}})\right \|^2_{\Omega_{\textsc{swot-nadir}}} \\
	+ & \lambda_3\left \| (X_{LR} + Y_{HR-cal} - Y_{\textsc{SAT}})\right \|^2_{\Omega_{\textsc{swot}}} 
	\end{split}
\label{eqn:calmapcost}
\end{equation}
with $||.||^2_{\Omega}$ the $l2$ norm computed on domain $\Omega$ and $\lambda_{1,2,3}$ weighing coefficients.

Regarding the prior term (\ref{eqn:4dprior}), we follow \cite{fablet_james} and consider a U-Net-like architecture for operator $\Phi$ to account for the multi-scale features of sea surface dynamics.

\subsection{End to end Learning}

Based on the variational formulation introduced in the previous section, we follow the general framework presented in \cite{fablet_james} to design an end-to-end architecture, which implements a trainable gradient-based solver for minimization (\ref{eqn:calmapcost}). This solver implements a given number $N$ of the following iterative update at iteration $k$:  
\begin{equation}
	\begin{split}
	& X^{(k+1)} =  X^{(k)} - \psi(\nabla_{X}[J(X^{(k)}, Y) + R(X^{(k)})])
	\end{split}
\end{equation}
with $\nabla_{x}$ the gradient operator w.r.t. the state $X$ derived using automatic differentiation tools. $\psi$ is parameterized as a recurrent LSTM network   \cite{lstm}.  The resulting end-to-end architecture uses as inputs an initial guess $X_{0}$
and a series of observation data $Y_{\textsc{SAT}}$. As initialization $X_{0}$, we consider the raw observation data where available.

The learning cost $L$ is decomposed as follows:
\begin{equation}
	\begin{split}
		& L = L_{4dVarNet} + L_{cal} \\
		& L_{cal} = \alpha_{\epsilon}||X_{cal} - \tilde{X}_{cal}||^2_{\Omega_{\textsc{swot}, t_c}} + \alpha_{\nabla\epsilon}||\epsilon_{\nabla cal}||^2_{\Omega_{\textsc{swot}, t_c}} \\
		& \epsilon_{\nabla cal} = ||\nabla X_{cal}|| - ||\nabla\tilde{X}_{cal}||
	\end{split}
	\label{eqn:loss}
\end{equation}
$L_{4dVarNet}$ is the supervised loss term described in \cite{e2e4dvar} comprised of a supervised reconstruction loss on the SSH fields and its gradient as well as regularization costs on $\Phi$  with an additional reconstruction cost on the low resolution estimate. $\tilde{X}_{cal}$ is the true SWOT SSH value interpolated on the target grid. $\nabla\epsilon_{cal}$ is the mse of the amplitude of the spatial gradients
The loss is only computed on the central time frame $t_c$ of the time window $\Delta T$.

\subsection{Training and implementation detail}

The models are trained using the Adam optimizer \cite{kingma2014adam} for approximately 200 epochs. During the training, the solver iterations increase progressively up to 15 and the learning rate decrease. 
We consider the state of 5 consecutive days to estimate the map of the central frame.
In order to compute the gradient of the variational cost, we leverage the automatic differentiation capabilities of the Pytorch library.
The interested reader can refer to our implementation\footnote{\url{https://github.com/CIA-Oceanix/4dvarnet-core/releases/tag/icassp2022}}.

\section{EXPERIMENTAL RESULTS}

\subsection{Data}
We run an Observing System Simulation Experiment  (OSSE) to evaluate the proposed framework.
We rely on NATL60 dataset, which refers to a realistic numerical simulation using NEMO model over the North Atlantic basin \cite{ajayi2020spatial}\cite{ajayi2021diagnosing}. 
In our the experiments, the case-study domain is a 10°x10° subpart of the GULFSTREAM area (33,43°N; -65,-55°W) and the target data is available as daily snapshots on a 1/10° spatial resolution grid.

We generate pseudo nadir altimeter observations sampled from realistic orbits using 2003 four-altimeter setting. We use SWOT simulator \cite{swot_simulator} to simulate realistic SWOT data which account for difference error sources, namely the roll, base dilatation, timing, phase errors and the karin noise. We refer the reader to 
\cite{swot_challenge}\cite{swot_simulator} for a more detailed presentation of this experimental setting. Besides altimeter data, we are also provided with the operational optimally-interpolated product from nadir altimeter data, referred to as DUACS. All these data are bilinearly interpolated on the target grid.

From the considered one-year simulation dataset, we take out 40 days in order to evaluate on the 20 middle days. This allows a 10-day buffer for the ocean to evolve between testing and training. The remaining 325 days are used for training and validation.

\subsection{Experimental setting}
We run three configurations of our framework, referred to as 4DVarNet-Calmap, 4DVarNet-Calmap${}_{\nabla}$ and 4DVarNet-Map, with different weighing factors  $\alpha_{\epsilon}$ and $\alpha_{\nabla\epsilon}$ defined in \ref{eqn:loss}. 
4dVarNet-Map with both $\alpha_{\epsilon}$ and $\alpha_{\nabla\epsilon}$ null can be regarded the proposed framework applied to the mapping of the SSH without making explicit the calibration of SWOT observation data.
Whereas 4dVarNet-Calmap with $\alpha_{\epsilon}=200$ ten times higher than $\alpha_{\nabla\epsilon}=20$ use the same ratio as  mapping loss $L_{4dVarNet}$, 
4dVarNet-Calmap${}_{\nabla}$ with $\alpha_{\epsilon}=20$ and $\alpha_{\nabla\epsilon}=200$ focuses on the reconstruction of the gradients on the swath.


For benchmarking purposes, we first consider  optimally-interpolated DUACS products: DUACS-4NADIRS issued
 from nadir altimeter data only and CAL+DUACS issued from nadir and noise-free SWOT data. The latter is regarded as an upper bound of the performance of this operational mapping when SWOT data will be available. Besides these operational SOTA products, we assess the performance of two deep learning schemes. 
We first apply a direct inverse model parameterized by  operator $\phi$ to directly output the targeted state from observation data. This baseline is labeled as Direct $\phi$. We also consider a SOTA deep learning architecture for computer vision and inpainting tasks, namely the vision transformer \cite{vit}\cite{inpaintingattn}. We may point out that the considered experimental setting involves much higher missing data rates, typically above 90\%, compared with typical inpainting case-studies \cite{inpaintingattn}.


\subsection{Results}

\begin{figure}[htb]
\begin{minipage}[b]{1.\linewidth}
  \centering
  \centerline{\includegraphics[width=8.0cm]{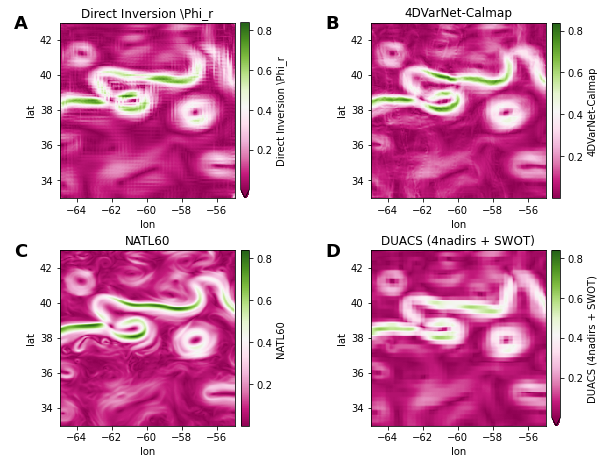}}
	\caption{Magnitude of the spatial gradient of SSH  }\smallskip
\end{minipage}
\end{figure}
%

In Table \ref{table:mapscores} we evaluate the benchmarked methods in terms of SSH mapping performance. We report the mean square error (MSE) of the reconstructed SSH maps w.r.t the ground truth along with the MSE for the amplitude of the spatial gradients of the SSH maps. We also assess the effective scale resolved in the reconstructed maps for each experiment as in \cite{bfn}. It comes to retrieving the smallest spatial wavelength for which the power spectral density of the true SSH is at least twice larger than that of reconstruction error. 


\begin{table}
{\footnotesize
\begin{tabular}{lrrr}
\toprule
metric &       mse  &  mse${}_\nabla$  & $\lambda_{er}$ \\
	&        ($1e^{-3}m^2$) &   ($1e^{-10}$) & (km) \\
\midrule
4DVarNet-Calmap${}_{\nabla}$ &  1.50 &  1.07 & 95.56 \\
\textbf{4DVarNet-Calmap}      &  \textbf{1.29} &  \textbf{1.03} & \textbf{88.21} \\
4DVarNet-Map           &  1.49 &  1.10 & 94.08 \\
DUACS 4NADIRS          &  2.53 &  1.95 & 129.26 \\
CAL + DUACS            &  2.20 &  1.74 & 125.32 \\
Direct $\phi$          &  2.10 &  1.66 & 116.58 \\
CAL + Direct ViT   &  2.48 &  2.04 & 126.02 \\
\bottomrule
\end{tabular}}
\caption{Mapping scores}
\label{table:mapscores}
\end{table}

\begin{figure}[htb]
\begin{minipage}[b]{1.\linewidth}
  \centering
  \centerline{\includegraphics[width=8.0cm]{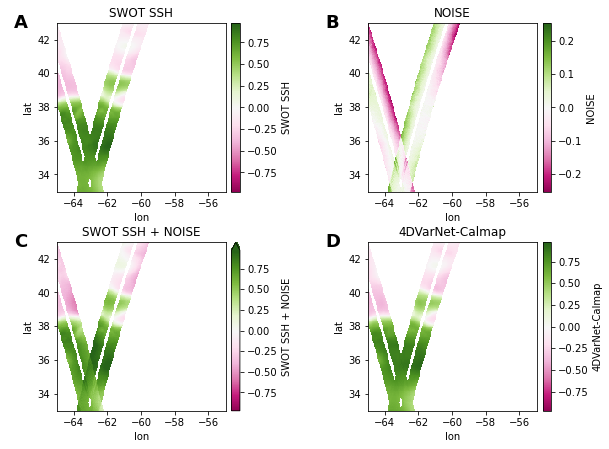}}
	\caption{(m) A: SWOT SSH signal, B: Acquisition noise, C: Noisy Observations, D: Calibrated SSH output of 4DVarNet-Calmap}
	\smallskip
\end{minipage}
%
\label{fig:res}

\end{figure}

In table \ref{table:mapscores} 4DVarNet-Calmap models clearly outperform the operational SOTA products.
Indeed we reduce the mse up to 41.6\% (resp. 40.0\%)  w.r.t. DUACS regarding the SSH (resp. is gradient amplitude). 
.
4DVarNet-Calmap schemes resolves finer scales from 30 to 36 km. 
If we compare the different parameterizations, we can first note that even without learning jointly the calibration, the 4DVarnet-Map framework shows remarkable robustness to the acquisition noise.
We may also point out that learning jointly the calibration improves the mapping performance if more focus  during the training stage is given to the reconstruction of the SSH on the swath.
The interest of the variational framework is emphasized by  the gain we get switching from a direct architecture to the variational formulation (Direct $\phi$ vs. 4dVarNet-Map).
Training a more complex model is non trivial because of the high rate of missing data in the maps to interpolate $>90\%$, as illustrated by the relatively poor performance of the Direct Vit.

In Table \ref{table:calscores}, we report the metrics of the different methods in the SWOT geometry: we interpolate again the estimated SSH on the swath. In order to handle ill-defined values that arise from this interpolation we do not consider the three outer most  across-track coordinates on each side of the nadir.
We compute the same MSE score as on the grid.
The effective resolution is computed for each pass of the satellite over the domain using the average wavenumber spectra in the along track direction. We report the mean and standard deviation of the scales resolved at each pass. The calibration metrics do benefit from the joint mapping and calibration. In order to recover finer scales of SSH signals, the focus has to be put on the reconstruction of the gradients during the training. 
\begin{table}
{\footnotesize
\begin{tabular}{lrrr}
\toprule
	{} &         mse &         mse${}_\nabla$ & $\lambda_{er}$  \\
	{} &    ($1e^{-4}$m²) &    ($1e^{-11}$) & (km) \\
\midrule
SWOT SSH + NOISE       &  51.23 &  39.16 & 51.37 ($\pm$ 56.21)\\
4DVarnet-Map           &  6.71 &  6.52 &  36.72 ($\pm$ 13.13) \\
4DVarnet-Calmap      &  \textbf{6.27} &  6.83 & 34.96 ($\pm$ 12.18)\\
\textbf{4DVarnet-Calmap}${}_{\nabla}$ &  6.79 &  \textbf{5.87} & \textbf{23.56} ($\pm$ \textbf{8.47}) \\
Direct  $\phi$        &  24.08 &  20.48 &  101.57 ($\pm$  35.23) \\
DUACS 4 NADIRS        &  27.34 &  22.50 &  97.21 ($\pm$ 85.90) \\
\bottomrule
\end{tabular}
}
\caption{Calibration scores}
\label{table:calscores}
\end{table}


\section{CONCLUSION}
We have investigated how data-driven methods can solve the joint calibration and mapping of noisy satellite altimetry data.
The proposed physics-informed learning scheme can truly benefit from the different data sources as well as prior physical knowledge. Using a simulation-based experiments, we have demonstrated its potential to outperform operational SOTA mapping methods with no requirement for a prior calibration. Future work will investigate how to transfer these results to real observation datasets.

\bibliographystyle{IEEEbib}
\bibliography{refs}

\end{document}